\documentclass[10pt,twocolumn,letterpaper]{article}

\usepackage{cvpr}
\usepackage{times}
\usepackage{epsfig}
\usepackage{graphicx}
\usepackage{amsmath}
\usepackage{amssymb}
\usepackage{url}

\usepackage[pagebackref=true,breaklinks=true,letterpaper=true,colorlinks,bookmarks=false]{hyperref}

\cvprfinalcopy 

\ifcvprfinal\fi
\begin{document}

\title{FoodX-251: A Dataset for Fine-grained Food Classification}

%
%
%

\author{Parneet Kaur\thanks{Part of the work done while an intern at SRI International.}
	\quad Karan Sikka$^{\vee}$\thanks{Corresponding author, karan.sikka@sri.com}
	\quad Weijun Wang$^{\ddagger}$
	\quad Serge Belongie$^{\amalg}$
	\quad Ajay Divakaran$^{\vee}$\\
	\\
$^{\vee}$SRI International, Princeton, NJ \\
$^\ddagger$Google, Los Angeles, CA \\
$^\amalg$Cornell Tech, New York, NY \\
}

\maketitle

\begin{abstract}
Food classification is a challenging problem due to the large number of categories, high visual similarity between different foods, as well as the lack of datasets for training state-of-the-art deep models. Solving this problem will require advances in both computer vision models as well as datasets for evaluating these models. In this paper we focus on the second aspect and introduce \emph{FoodX-251}, a dataset of $251$ fine-grained food categories with $158k$ images collected from the web. We use $118k$ images as a training set and provide human verified labels for $40k$ images that can be used for validation and testing. In this work, we outline the procedure of creating this dataset and provide relevant baselines with deep learning models. The FoodX-251 dataset has been used for organizing iFood-2019 challenge\footnote{https://www.kaggle.com/c/ifood-2019-fgvc6} in the Fine-Grained Visual Categorization workshop (FGVC6 at CVPR 2019) and is available for download.\footnote{https://github.com/karansikka1/iFood\_2019}
\end{abstract}

\section{Introduction}
\label{sec:intro}

A massive increase in the use of smartphones has generated interest in developing tools for monitoring food intake and
trends~\cite{puri2009recognition, zhang2015snap, Mey2015}. Being able to estimate calorie intake can aid users to modify
their food habits and maintain a healthy diet. Current food journaling applications such as Fitbit App~\cite{fitbit}, MyFitnessPal~\cite{myfitnesspal} and My Diet Coach~\cite{dietcoach} require users to enter their meal information manually. A study of 141 participants in~\cite{cordeiro2015rethinking} reports that $25\%$ of the participants stopped food journaling because of the effort involved while $16\%$ stopped because they found it to be time consuming. On the other hand, designing a computer vision based solution to measure calories from clicked images would make the process very convenient. Such an algorithm would generally be required to solve several sub-problems $-$ classify, segment and estimate 3D volume of the given food items. Our focus in this work is to provide a dataset to facilitate the first task of classifying food items in still images. 

Food classification is a challenging task due to several reasons: large number of food categories that are fine-grained in nature, resulting in high intra-class variability and low inter-class variability (\eg, different varieties of pasta), prevalence of non-rigid objects, and high overlap in food item composition across multiple food dishes. Further, in comparison to standard computer vision problems such as object detection~\cite{lin2014microsoft} and scene classification~\cite{zhou2017places}, the datasets for food classification are limited in both quantity and quality to train and evaluate deep neural networks. In this work we push the current research in food classification by introducing a new dataset of $251$ fine-grained classes with $158k$ images that supersedes prior datasets in number of classes and data samples.

\begin{figure}[]
    \centering
	\includegraphics[width = 3in]{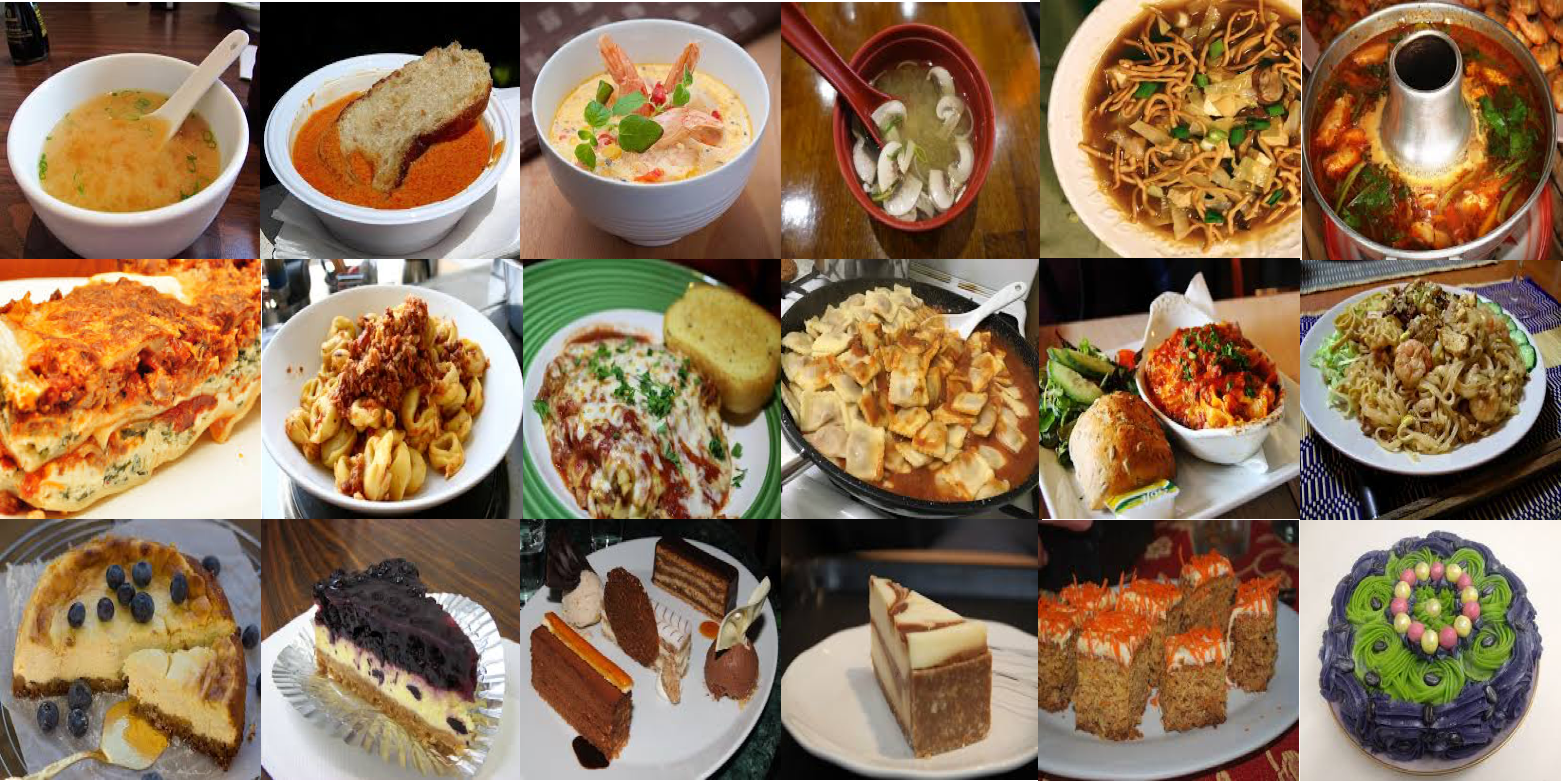}
	\caption{FoodX-251 Dataset. We introduce a new dataset of $251$ fine-grained classes with $118k$ training, $12k$ validation and $28k$ test images. Human verified labels are made available for the training and test images. The classes are fine-grained and visually similar, for example, different types of cakes, sandwiches, puddings, soups, and pastas.}
	\label{fig:intro}
\end{figure}

\section{Related Work}

\begin{figure*}[!t]
    \centering
	\includegraphics[width = 5.5in]{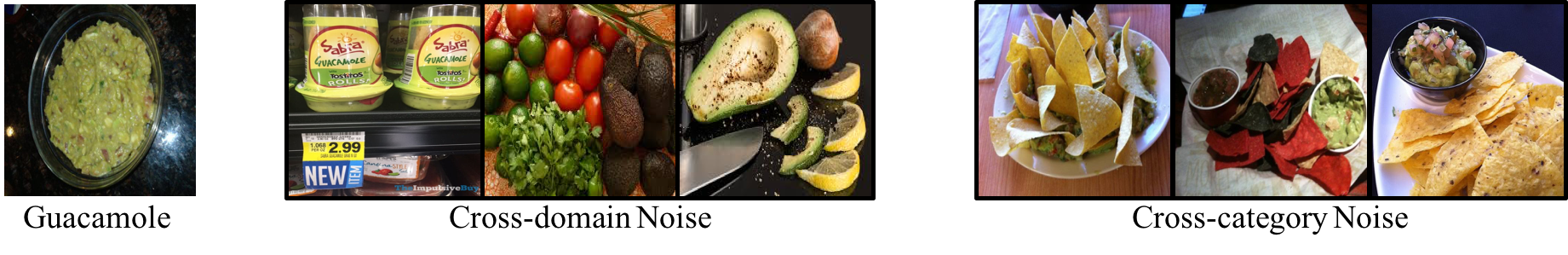}
	\caption{Noise in web data. \emph{Cross-domain noise}: Along with the images of specific food class, web image search also includes images of processed and packaged food items and their ingredients. \emph{Cross-category noise}: An image may have multiple food items but only one label as its ground truth.
	}
	\label{fig:noise}
\end{figure*}

\begin{table*}[!h]
	\centering
	\begin{tabular}{|c|c|c|c|c|}
		\hline
		\textbf{Dataset} & \textbf{Classes} & \textbf{Total Images} & \textbf{Source} & \textbf{Food-type} \\ \hline
		ETHZ Food-101~\cite{Bos2014} & 101 & 101,000 & foodspotting.com & Misc. \\ \hline
		UPMC Food-101~\cite{wang2015recipe} & 101 & 90,840 & Web  & Misc.\\ \hline
		Food50~\cite{joutou2009food} & 50 & 5000 & Web  & Misc.\\ \hline
		Food85~\cite{hoashi2010image} & 85 & 8500 & Web  & Misc.\\ \hline
		CHO-Diabetes~\cite{anthimopoulos2013segmentation} & 6 & ~5000 & Web  & Misc.\\ \hline
		Bettadapura et al.~\cite{bettadapura2015leveraging} &  75 & 4350  & Web, smartphone  & Misc. \\\hline
		UEC256~\cite{Kaw2014} & 256 & at least $100$ per class & Web & Japanese\\ \hline
		ChineseFoodNet~\cite{chen2017chinesefoodnet} & 208 & 185,628 & Web & Chinese\\ \hline		
		NutriNet dataset~\cite{mezgec2017nutrinet} &  520 & 225,953  & Web  & Central European \\\hline
		\textbf{Food-251 }& \textbf{251} & \textbf{158,846} & \textbf{Web} & \textbf{Misc.} \\ \hline
	\end{tabular}
	\caption{Datasets for food recognition. In comparison to prior work, the FoodX-251 dataset (1) provides more classes and images than existing datasets and (2) features miscellaneous classes as opposed to a specific cuisine/food type.}
\label{tbl:food_datasets}
\end{table*}

Earlier works have tried to tackle the issue of limited datasets for food classification by collecting training data
using human annotators or crowd-sourcing platforms~\cite{farinella2016retrieval,chen2012automatic,Kaw2014,zhang2015snap,
Mey2015}. Such data curation is expensive and limits the scalability in terms of number of training categories as well
as number of training samples per category. Moreover, it is challenging to label images for food classification tasks as
they often have co-occurring food items, partially occluded food items, and large variability in scale and viewpoints.
Accurate annotation of these images would require bounding boxes, making data curation even more time and cost
prohibitive. Thus, it is important to build food datasets with minimal data curation so that they can be scaled to novel
categories based on the final application. Our solution is motivated by recent advances in exploiting the knowledge available in web-search engines and using it to collect a large-scale dataset with minimal supervision~\cite{kaur2017combining}.

Unlike data obtained by human supervision, web data is freely available in abundance but contains different types of noise~\cite{chen2015webly, wang2008annotating, sukhbaatar2014learning}. Web images collected via search engines may include images of processed and packaged food items as well as ingredients required to prepare the food items as shown in Figure~\ref{fig:noise}. We refer to this noise as cross-domain noise as it is introduced by the bias due to specific search engine and user tags. In addition, the web data may also include images with multiple food items while being labeled for a single food category (cross-category noise). For example, in images labeled as Guacamole, Nachos can be predominant (Figure~\ref{fig:noise}). Further, the web results may also include images not belonging to any particular class.

Table~\ref{tbl:food_datasets} lists prior datasets for food classification. ETHZ Food-101~\cite{Bos2014} consists of $101,000$ images of $101$ categories. The images are downloaded from a photo sharing website for food items (foodspotting.com). The test data was manually cleaned by the authors whereas the training data consists of cross-category noise, \ie, images with multiple food items labeled with a single class. UPMC Food-101~\cite{wang2015recipe} consists of $90,840$ images for the same $101$ categories as ETHZ Food-101 but the images are downloaded using web search engine. 
Some other food recognition datasets with fewer food categories~\cite{joutou2009food, hoashi2010image, anthimopoulos2013segmentation, bettadapura2015leveraging} are also listed in Table~\ref{tbl:food_datasets}. In comparison to these datasets, our dataset consists of more classes $(251)$ and images $(158k)$.

UEC256~\cite{Kaw2014} consists of $256$ categories with bounding box indicating the location of its category label. However, it mostly contains Japanese food items. ChineseFoodNet~\cite{chen2017chinesefoodnet} consists of $185,628$ images from $208$ categories but is restricted to Chinese food items only. NutriNet dataset~\cite{mezgec2017nutrinet} contains $225,953$ images from $520$ food and drink classes but is limited to Central European food items. In comparison to these datasets, out dataset consists of miscellaneous food items from various cuisines.

\section{FoodX-251 Dataset}
We introduce a new dataset of $251$ fine-grained (prepared) food categories with $158k$ images collected from the web. We provide a training set of $118k$ images and human verified labels for both the validation set of $12k$ images and the test set of $28k$ images. The classes are fine-grained and visually similar, for example, different types of cakes, sandwiches, puddings, soups and pastas.

\subsection{Data Collection}
We start with the $101$ food categories in Food-101 dataset~\cite{Bos2014} and extract their sibling categories from WordNet~\cite{miller1995wordnet,bird2009natural}. We first manually filter and remove all non-food or ambiguous classes.\footnote{By ambiguous we refer to those food classes where people do not seem to have a visual consensus.} Since our primary aim is fine-grained food classification task, we also remove general food classes. For example, different types of pastas and cakes are included but ``pasta'' and ``cake'' are removed from the list. This gives us $251$ food classes. 

For each class, we use web image search to download the corresponding images. Due to the nature of images on these search engines, these images often include images of processed and packaged food items and their ingredients resulting in cross-domain noise. We also observe cross-category noise when for a image search with a single food-item, some images that have multiple food items are downloaded (see Figure ~\ref{fig:noise}).

\begin{figure}[!t]
    \centering
	\includegraphics[width = 3in]{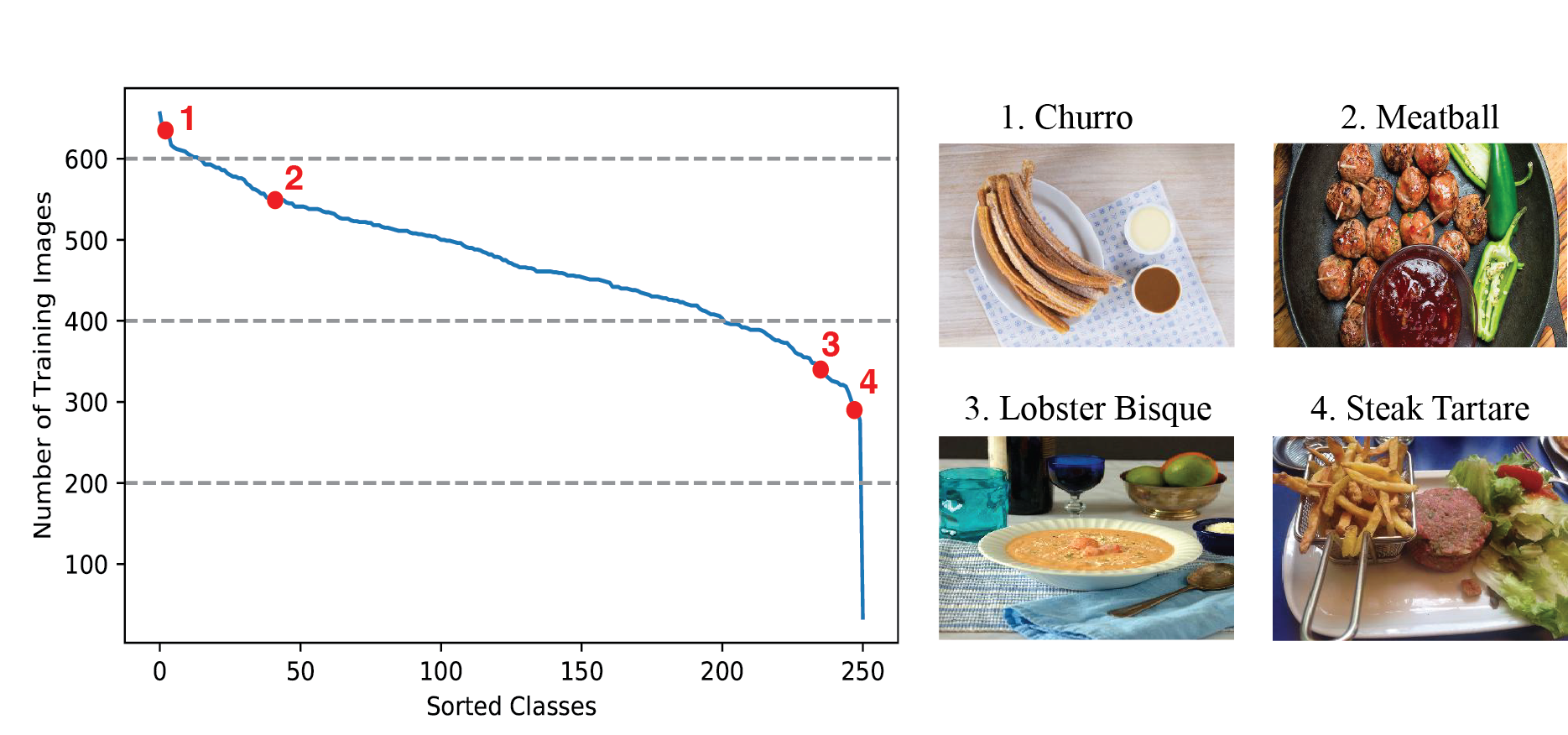}
	\caption{[Left] Training images distribution per class. [Right] Representative images for 4 sampled classes. Food items such as ``churro'' and ``meatball'' have large numbers of training images while food items such as ``lobster bisque'' and ``steak-tartare'' have relatively fewer training images.}
	\label{fig:distribution}
\end{figure}

We further filter exact as well as near-exact duplicate images from the dataset. We then randomly selected $200$ images from each class and have human raters (3 replications) do verification on this set. From the verified set, we randomly select $70\%$ images for testing and $30\%$ for validation. We use all the remaining images as the training set. The human verification step ensures that the validation and the test set are clean of any cross-domain or cross-category noise. Example of categories with large numbers of samples are generally popular food items such as 
``churro'' or ``meatball,'' while examples of categories with lower numbers of samples are less popular items such ``marble cake,'' ``lobster bisque,'' and ``steak-tartare''~(Figure~\ref{fig:distribution}).  


\subsection{Evaluation Metric}
We follow a similar metric to the classification tasks of the ILSVRC \cite{deng2012ilsvrc}. For each image $i$, an algorithm will produce $3$ labels $l_{ij},j=1,2,3$, and has one ground truth label $g_i$. The error for that image is:

\begin{equation}
e_i = \min_j d(l_{i,j},g_i),
\end{equation}

where,

\begin{equation}
d(x,y) = \begin{cases}
0, & \text{if $x=y$}.\\
1, & \text{otherwise}.
\end{cases}
\end{equation}

The overall error score for an algorithm is the average error over all $N$ test images:
\begin{equation}
score = \frac{1}{N} \sum_i e_i.
\end{equation}

\begin{table}[!h]
    \centering
    \begin{tabular}{|c||c|c|c|} \hline
         Method & \multicolumn{3}{|c|}{Top-3 Error $\%$}  \\
         & Val. & \multicolumn{2}{|c|}{Test} \\ 
         & & Public & Private \\ \hline
         ResNet-101 & $0.36$ & $0.37$ & $0.37$ \\ 
         (\emph{finetune} last-layer) &  & &  \\ \hline
         ResNet-101 & $0.16$ & $0.17$ & $0.17$ \\
         (finetune all-layers) &  & &  \\ \hline
    \end{tabular}
    \vspace{1em}
    \caption{Table reports the baseline performance on the FoodX-251 dataset on the validation and the test set.
    }
    \label{tab:baseline}
\end{table}

\subsection{Baseline Performance}
We implement a naive baseline using a pre-trained ResNet-101 network \cite{he2016deep}. We train the model using ADAM optimizer \cite{kingma2014adam} with a learning rate of $5e^{-5}$, which is dropped by a factor of $10$ after every $10$ epochs. The model is trained for a maximum of $50$ epochs with early stopping criteria based on the performance on the validation set. We use random horizontal flips and crops for data augmentation. We use the model checkpoint with the best performance on the validation set for computing test set performance. We have shown results for the validation splits and test splits (as per the Kaggle challenge page) in Table \ref{tab:baseline}.  

We observe that ResNet-101 model fine-tuning only the last layer shows a significantly lower performance as compared to the model with fine-tuning all the layers ($0.37$ vs. $0.17$ respectively). We believe that this occurs since the original pre-trained filters are not well suited to the food classification task. As a result, fine-tuning the entire network helps in improving the performance on the fine-grained classification task by a noticeable margin.

\section{iFood Challenge at FGVC workshop}
The FoodX-211 dataset was used in the iFood-2019 challenge\footnote{https://www.kaggle.com/c/ifood-2019-fgvc6} in Fine-Grained Visual Categorization workshop at CVPR 2019 (FGVC6).\footnote{https://sites.google.com/view/fgvc6} The dataset is also available for download.\footnote{https://github.com/karansikka1/iFood\_2019}

This dataset is an extension of FoodX-211 dataset which was used to host iFood-2018 challenge\footnote{https://github.com/karansikka1/Foodx} at FGCV5 (CVPR 2018). FoodX-211 had 211 classes with $101k$ training images, $10k$ validation images and $24k$ test images.

\section{Conclusions}
In this work, we compiled a new dataset of food images with $251$ classes and $158k$ images. We also provide human-verified labels for $40k$ images. The baseline results using state-of-the-art ResNet-101 classifier shows $17\%$ top-3 error rate. There is an opportunity for the research community to use more sophisticated approaches on this dataset to further improve the classifier performance. We hope that this dataset will provide an opportunity to develop methods for automated food classification as well as serve as a unique dataset for the computer vision research community to explore fine-grained visual categorization.

\section{Acknowledgements}
We are thankful to the FGVC workshop organizers for the opportunity to host the iFood competition. We gratefully acknowledge SRI International for providing resources for data collection and Google for providing resources for labeling the data. We are also thankful to Tsung-Yi Lin and CVDF for helping with uploading the data, and also Maggie Demkin, Elizabeth Park, and Wendy Kan from Kaggle for helping us set up the challenge.

{\small
\bibliographystyle{ieee_fullname}
\bibliography{egbib}
}

\end{document}